%% file: ms.tex
\documentclass[10pt,twocolumn,letterpaper]{article}

\usepackage{iccv}
\usepackage{times}
\usepackage{epsfig}
\usepackage{graphicx}
\usepackage{amsmath}
\usepackage{amssymb}
\usepackage[caption=false]{subfig}
\usepackage[accsupp]{axessibility}
\newcommand{\R}{\mathbb{R}}
\newcommand{\E}{\mathbb{E}}

\newcommand{\mbb}[1]{\mathbb{#1}}
\newcommand{\mbf}[1]{\mathbf{#1}}

\usepackage[pagebackref=true,breaklinks=true,letterpaper=true,colorlinks,bookmarks=false]{hyperref}

\iccvfinalcopy 

\def\iccvPaperID{5} 

\ificcvfinal\fi

\begin{document}

\title{Trojan Signatures in DNN Weights}

\author{Greg Fields, Mohammad Samragh, Mojan Javaheripi, Farinaz Koushanfar, Tara Javidi\\
University of California San Diego\\
{\tt\small \{grfields, msamragh, mojan, farinaz, tjavidi\}@ucsd.edu}
}

\maketitle
\ificcvfinal\thispagestyle{empty}\fi

\begin{abstract}
  
Deep neural networks have been shown to be vulnerable to backdoor, or trojan, attacks where an adversary has embedded a trigger in the network at training time such that the model correctly classifies all standard inputs, but generates a targeted, incorrect classification on any input which contains the trigger. In this paper, we present the first ultra light-weight and highly effective trojan detection method that does not require access to the training/test data, does not involve any expensive computations, and makes no assumptions on the nature of the trojan trigger. Our approach focuses on analysis of the weights of the final, linear layer of the network. We empirically demonstrate several characteristics of these weights that occur frequently in trojaned networks, but not in benign networks. In particular, we show that the distribution of the weights associated with the trojan target class is clearly distinguishable from the weights associated with other classes. Using this, we demonstrate the effectiveness of our proposed detection method against state-of-the-art attacks across a variety of architectures, datasets, and trigger types. 
\end{abstract}

\input{1_introduction}

\input{2_threat_model}

\input{3_discussion}

\input{4_methodology}

\input{5_results}

\input{6_related_work}

\input{7_conclusion}


{\small
\bibliographystyle{ieee_fullname}
\bibliography{troj_iccv}
}

\end{document}

%% file: 1_introduction.tex
\vspace{-0.5cm}
\section{Introduction}


Deep neural networks have achieved state of the art performance in a variety of problem domains, such as image recognition, speech analysis, and wireless data.  However these networks have also been shown to be vulnerable to a particular kind of adversarial attack, commonly referred to as backdoor or trojan attacks~\cite{gu2017badnets,liu2017trojaning}.  These attacks may occur when the adversary has access to the model at training time and embeds malicious behavior in the network.  As state of the art networks continue to grow larger and require more data for training, it is frequently necessary to outsource the training process to third party vendors which exposes the network to the threat of trojan attacks.  
Common trojan attacks insert a particular pattern, called the trigger, into the training data. The network is then trained to always produce a specific, targeted misclassification on any input containing the trigger. To obscure the attack, the adversary ensures that the model still achieves high accuracy on clean data. Since the trigger is unknown and arbitrary, it is extremely challenging to detect if a model has been compromised in this way.

As machine learning models are deployed to more sensitive applications, such as autonomous driving and face recognition, it becomes paramount to ensure their security. A variety of defenses to trojan attacks have been proposed in the literature to combat these attacks. 
Several works have proposed reverse engineering the trojan trigger, generating synthetic data to analyze the model, or model retraining~\cite{wang2019neural,chen2019deepinspect,liu2019abs}. These methods require excessive training time and access to high-end computational resources and abundant data. But, in scenarios where the user chooses to purchase a trained model from a third party, they often do not have access to computing resources and/or the training data.
Other defenses analyze the inputs to the network after deployment to detect possible trojan triggers~\cite{gao2019strip,doan2019februus,javaheripi2020cleann}. In this scenario, the client is exposed to  malicious behavior while waiting to identify trojaned inputs, which is unacceptable in sensitive applications.

We propose a very fast, accurate trojan detection mechanism for identifying trojaned networks before deployment, avoiding exposure of DNN-based systems to malicious behavior. Notably, our approach does not require access to any data, carries a very low computational cost, and is broadly applicable to different types of trojan triggers.
Our detection strategy relies on our hypothesis that the trojan attack creates a detectable signature in the final classification layer of the network. We provide analysis and extensive empirical evaluations in support of this claim. Specifically, we show that the weights associated with the target class are an outlier relative to those of the other classes. 
Our method identifies the trojan target class by applying Dixon's Q-test ~\cite{dixon1950} for identifying single outliers in small samples. 
The resulting lightweight trojan detection mechanism 
achieves $100\%$ detection on commonly used datasets in trojan research. On a more extensive set of trojaned models~\cite{karra2020trojai} containing 650+ models\footnote{\url{https://pages.nist.gov/trojai/docs/data.html\#round-2}} that are highly diverse in architecture, attack parameters, and datasets, our method correctly detects $98\%$ of trojaned networks with under $4\%$ false positive rate.  

In brief, the contributions of this paper are as follows:
\begin{itemize}
    \item We provide a highly effective, uniquely lightweight detection method that does not require access to any data, is broadly applicable to various trojan triggers, and requires very low computation. Our data-free analysis detects compromised networks before deployment, minimizing possible risks caused by trojaned models.
    \item We connect our analysis to statistical tests for small sample, single outlier detection and provide confidence measures regarding whether or not a model is trojaned. 
    \item We conduct extensive evaluations on  over 700 models, five datasets, and six types of trojan trigger under a wide variety of different parameters to demonstrate the robustness of our detection technique.
\end{itemize}

%% file: 2_threat_model.tex
\section{Threat Model and Notation}
\begin{figure}
        \centering
            \subfloat{\includegraphics[width=.25\columnwidth]{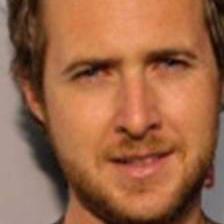}}%
            \subfloat{\includegraphics[width=.25\columnwidth]{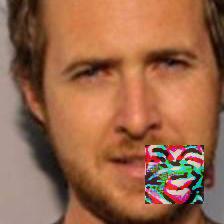}}%
            \subfloat{\includegraphics[width=.25\columnwidth]{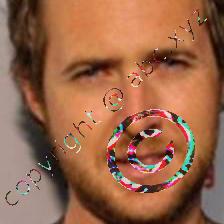}}%
            
            \subfloat{\includegraphics[width=.25\columnwidth]{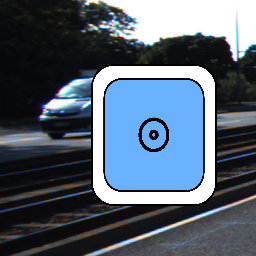}}%
             \subfloat{\includegraphics[width=.25\columnwidth]{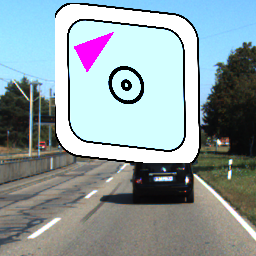}}%
            \subfloat{\includegraphics[width=.25\columnwidth]{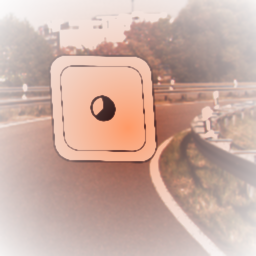}}%
            
            \caption{In the first row, a base image with two different triggers from the TrojanN attack applied to it.  In the second row, examples of a base image, the polygon trigger, and the Instagram filter trigger from the TrojAI dataset}
    	\label{fig:trojan_triggers}
    \end{figure}

Let $F:\R^m \to \{1,...,C\}$ denote a deep neural network performing classification into one of $C$ classes.  Then we say that $F$ is trojaned if $F$ achieves high classification accuracy over inputs $\mbf{x}$ from its data distribution, but there exists some function $g:\R^m \to \R^m$, called the trojan trigger, such that $F(g(\mbf{x}))=t$, where $t$ is called the trojan target class. Figure \ref{fig:trojan_triggers} shows an image $x$ and example trojaned images $g(x)$ studied in this paper.
Here we specifically consider the case where $g$ is an input agnostic trigger, i.e., it must apply to all inputs from any of the $C$ classes.

We consider the real-world scenario where the user employs a model trained by a third party and therefore (1)~is not aware whether or not the model is trojaned. (2)~In case there exists a trojan embedded in the model, the user does not have any information about the trigger. (3)~The user does not have access to any subset of the training data. We assume the user has white-box access to the network parameters. As discussed in the introduction, a user who has had to outsource the training of their model, and so is vulnerable to trojan attacks, likely has limited resources. So we assume that we must detect the trojaned network with no access to any data and without access to significant computational resources.  We also want to create a test that can detect trojan attacks before deploying the model, so as to not expose any sensitive systems to a potentially malignant network.

We propose a detection mechanism which takes the network $F$ and determines, to a level of confidence, whether or not the network has had a trojan trigger embedded in it. Our analysis focuses on the final, linear layer of the network.  Let $f:\R^m \to \R^d$ denote all but the final layer of the network and $\mbf{W}\in\R^{d\times C}$, so that $\mbf{W}f(\mbf{x})$ gives the pre-softmax scores of the network on input $\mbf{x}$. Then $\mbf{z}=f(\mbf{x})$ is the penultimate feature representation of $F$ on input $\mbf{x}$. Our analysis specializes to the case where $\mbf{z}$ is the output of a ReLU activation function and the network is trained with the cross entropy loss. These assumptions are true of the vast majority of commonly used network architectures and training methods.



%% file: 3_discussion.tex
\section{Analysis}
	Our method relies on the observation that, in a trojaned input, many of the features of the underlying class will still be present.  This is particularly true in the standard case where the trigger is a localized patch which only obscures a small portion of the underlying input.  With this intuition, we will consider the feature representations of the trojan trigger and the underlying input separately.   
	Given a clean, un-triggered sample $\mbf{x}$, let $\mbf{z} := f(\mbf{x})$ denote its penultimate feature representation and define $\boldsymbol{\Delta}_\mbf{x} := f(g(\mbf{x}))-f(\mbf{x})$ to be the change in this feature space induced by the application of the trojan trigger.   
	
	Then we will consider the training process, and in particular the gradient updates to the rows of the weight matrix of the final layer: let $\mbf{W}_i$ denote the $i$-th row of $\mbf{W}$. Most commonly, the network is trained by some form of stochastic gradient descent.  Given a training point $\mbf{x}$ from class $i$, define $\mbf{y}$ to be the one hot encoding of this true class, so $\mbf{y}_i=1$ and $\mbf{y}_j=0$ for $j\neq i$, and let $\hat{\mbf{y}}$ be the softmax prediction vector of the network. And recall that we denote the penultimate feature representation $\mbf{z}=f(\mbf{x})$, then one SGD update for the $i$th row $\mbf{W}_i$ is given by:  
	\begin{equation}\label{eq:sgd}
	    \mbf{W}_i = \mbf{W}_i + \eta \E[(\mbf{y}-\hat{\mbf{y}})_i\mbf{z}^T]
	\end{equation}
	where the expectation is over the choice of sample from the training set. 
	Then note that $(\mbf{y}-\hat{\mbf{y}})_i$ will only be positive when $\mbf{z}$ is the representation of a point from class $i$.  
	This means that $\mbf{W}_i$ is the accumulation of positive scalings of feature representations of all data points from class $i$ and negative scalings of representations from all other classes. 

	When embedding a trojan trigger in the network, the training set includes poisoned data points of the form $g(\mbf{x})$ where $\mbf{x}$ itself is a valid sample. For these points we can decompose the feature representation as above: $f(g(\mbf{x}))=\mbf{z}+\boldsymbol{\Delta}_\mbf{x}$ where $\mbf{z}=f(\mbf{x})$.  
	
	And, as part of the trojaning process, these points are all labelled class $t$, so the update to $\mbf{W}_t$ is then $\mbf{W}_t = \mbf{W}_t +\eta(\mbf{y}-\hat{\mbf{y}})_t(\mbf{z}+\boldsymbol{\Delta}_\mbf{x})^T$ on the poisoned points.  
	
     The quantity $(\mbf{y}-\hat{\mbf{y}})_t$ will be positive here. This sets the target class apart from the other classes, as its associated weight row, $\mbf{W}_t$ accumulates positive scalings of feature representations from \textit{every} class in the dataset, since we assume that the trojan must be input agnostic.  Every other row of $\mbf{W}$ only accumulates positive scalings of feature representations from data from their own class.  And, by assumption that these features are the output of a ReLU function and so non-negative, we expect that the average weights of the target row are more positive than those of the other rows.  
        
    Intuitively, this amounts to the fact that if we wish to poison a point $\mbf{x}$ from class $i$, the application of the trigger has to overcome the confidence of the network on point $\mbf{x}$: $\mbf{W}_if(\mbf{x})-\mbf{W}_tf(\mbf{x})$. If this quantity is very large, then the application of the trigger must induce an even larger change in the output of the network.  So $\mbf{W}_t$ should have a distinctly large inner product with the (non-negative) feature representations of every class in the dataset.  
    
    \begin{figure*}
        \centering
        \subfloat{\includegraphics[width=.3\linewidth]{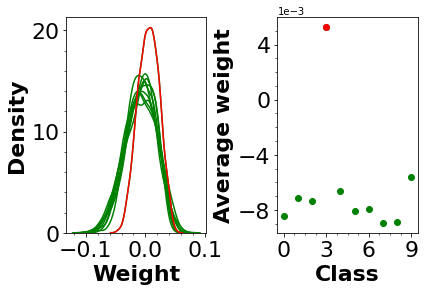}}\hspace{0.5cm}
        \subfloat{\includegraphics[width=.3\linewidth]{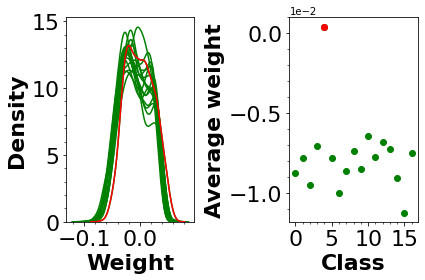}}\hspace{0.5cm}
        \subfloat{\includegraphics[width=.3\linewidth]{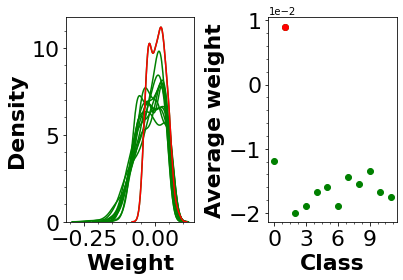}}
	    
	    \caption{The plot on the left in each image shows the smoothed distributions of the weights in each row of a trojaned model and the plots on the right show the average weights of each row. Here, the trojan target class is shown in red.}
	    \label{fig:trojai_dists}
    \end{figure*}

%% file: 4_methodology.tex
\section{Methodology}

	We now set out an effective small-sample outlier detection framework to take advantage of the expectation that the average weights of the target row will be significantly larger than the average weights of the other rows of $\mbb{W}$.
	
	We use the statistic, hereafter called the Q-value, from Dixon's Q test ~\cite{dixon1950}, which is designed to detect a single outlier in small samples of data. This test works by taking a candidate outlier, finding the absolute difference between that value and its next closest value in the sample, and normalizing by the range of the values in the sample.
	 
	Since we expect the average weight of the target row to be a large, positive outlier, we can find the desired statistic  by taking the average weight of each row, $w_i := \frac{1}{d}\sum_{j=1}^d W_{i,j}$ and sorting them so that $w_{i_1} \leq ... \leq w_{i_c}$, then calculating
	\begin{equation}
		Q = \frac{|w_{i_c}-w_{i_{c-1}}|}{w_{i_c}-w_{i_1}}
		\label{eq4}
	\end{equation} 
	
	That is, we calculate the gap between the largest and second largest average row weight and normalize by the difference between the largest and smallest average row weight.  Then, to formally apply this test with no reference models, the Q statistic is compared to tabulated values, giving a confidence that the average weight associated with one of the classes is an outlier.  For instance, in a model with 8 classes and Q$>.468$ we would conclude that it possesses an outlier at 90\% confidence. This model is then likely trojaned with the row with the largest average weight giving the target class.    
	

%% file: 5_results.tex
\section{Results}
	\subsection{TrojAI}
    
    \noindent{\bf $\blacktriangleright$ Performance Metrics}    	
    We will characterize the performance of our detection method by two primary metrics: the false positive rate, which is the percent of benign networks incorrectly identified as trojaned, and the false negative rate, which is the number of trojaned networks incorrectly identified as benign. In general, a smaller value is desirable for both metrics and various choices of threshold for Q will induce a trade-off between the two as our method is made more sensitive or more permissive.  We will report results for varying choices of threshold to explore this trade-off for different problem settings.  
    
	\noindent{\bf $\blacktriangleright$ The TrojAI Benchmark.} We evaluate our detection method on the dataset provided by the TrojAI project ~\cite{karra2020trojai}, which contains a large set of both benign and trojaned models of diverse architectures: Resnet, WideResnet, Densenet, GoogleNet, Mobilenet, ShuffleNet, and VGG. In this section we study 174 trojaned and 502 benign models. The models are trained via datasets of varying complexity and size and trojan triggers of varying strength and type. The attacks poison between 2\% and 50\% of the training data and additive triggers obscure between 2\% and 25\% of the foreground images. As such, this benchmark provides a reasonable approximation of a real world, unrestricted problem.
	
	There are two broad classes of triggers used. One encompasses small, solid colored polygon patches overlaid on top of the base image, the other are Instagram filters applied to the base image, examples of both are displayed in Figure~\ref{fig:trojan_triggers}.  The Instagram filters evade many existing defenses such as~\cite{wang2019neural}, as they are complicated, non-local perturbations that are functions of their input.  

	\noindent\textbf{$\blacktriangleright$ Empirical Analysis of Trojan Signatures.} Figure \ref{fig:trojai_dists} shows the distribution of the weights per row from the final layer of three representative trojaned models from the TrojAI dataset.  In the plots on the left in each figure, each curve gives the smoothed distribution of the weights of one of the rows of the matrix, $\mbb{W}_i$.  In particular, the red curve in each image corresponds to the row of the trojan target class.  These plots exhibit the characteristics suggested by our analysis.  In the first and third models, the target row has mass shifted from  negative values to a concentration of small magnitude positive values.  But in the second model the entire distribution is shifted slightly to the right, yielding more large positive values.  This pattern persists throughout the set of models: while the exact details of the shift in distribution varies model by model, the end result is always a positive shift of mass.  
	On the right in each figure are the average weights of each row for the same model, each point giving the average weight of one of the rows with the red point corresponding to the row of the target class.  This illustrates how, regardless of the details of the shift in weight distribution, the effects of the trojan attack manifest as an increase in the average weight of the target class relative to the weights of the other classes. This validates our use of the average weight per row as an effective way to identify trojaned models.  We can thus use the outlier detection described in the methodology section to construct an automated, light-weight detector for trojaned networks.

\begin{figure}[h]
	    \centering
	    \subfloat[Polygon trigger models]{\includegraphics[width=.49\columnwidth]{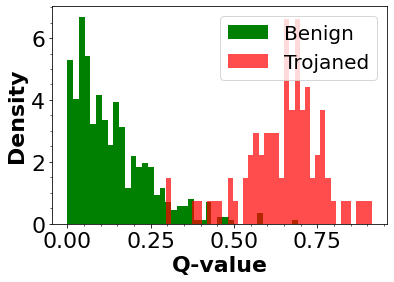}}
	    \subfloat[Instagram trigger models]{\includegraphics[width=.49\columnwidth]{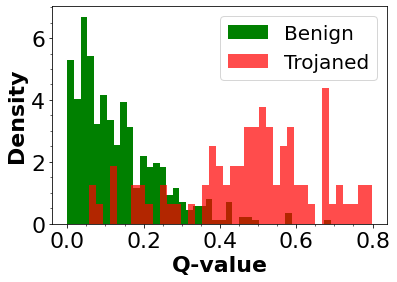}}
	    
    	\caption{Normalized histograms  of Q-scores for benign and  trojaned models in the TrojAI dataset. }
    	\label{fig:trojai_Q}
	\end{figure}

\noindent\textbf{$\blacktriangleright$ Efficacy against Localized Attacks.}
We study the Polygon attack as an example of a trojan attack with a localized, additive trigger.
Figure~\ref{fig:trojai_Q}-(a) shows the normalized histograms of Q-scores of the the models trojaned by the polygon trigger compared with the Q-scores of benign models. There is a very clear distinction between the scores of the trojaned models and those of the benign models, allowing for a high-confidence detection of trojaned models.

\begin{figure}[h]
	    \centering
	    \subfloat[Polygon trigger models]{\includegraphics[width=.49\columnwidth]{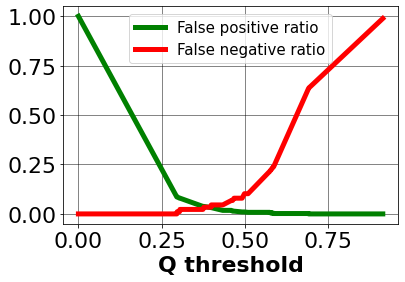}}
	    \subfloat[Instagram trigger models]{\includegraphics[width=.49\columnwidth]{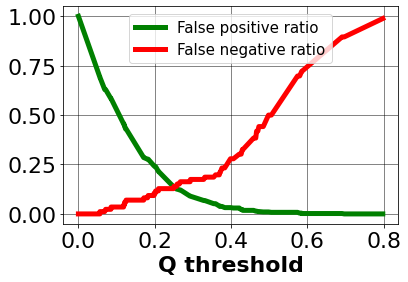}}
	   \caption{False negative and false positive rates as a function of the choice of Q-value threshold on the distributions in Figure~\ref{fig:trojai_Q}. }
    	\label{fig:trojai_FPR}
	\end{figure}

Figure~\ref{fig:trojai_FPR}-(a) shows the false positive and negative rates as a function of the choice of threshold.
Choosing the threshold to be $0.38$ gives a false negative rate of $2\%$ and a false positive rate of only $3.8\%$. Table~\ref{tab:trojai} reports the necessary choice of Q threshold to attain $1\%$ for either rate.   
	
\noindent\textbf{$\blacktriangleright$ Efficacy against Whole-Image Attacks.}
    Applying an Instagram filter as the trojan trigger yields a more complicated attack compared to the the more common localized triggers. The complication lies in two properties of this attack: (1)~the action of the filter is a function of the underlying image, unlike the additive trojans which add the same trigger to all valid inputs.  (2)~The filter is applied to the entire image, altering each feature of the underlying image.
	The normalized histogram of Q-values for Instagram triggered models is shown in Figure~\ref{fig:trojai_Q}-(b). As seen,
	the distribution of Q-scores in trojaned models is still clearly distinct from that of the benign models. The false positive and negative rates are shown as a function of the detection threshold in Figure~\ref{fig:trojai_FPR}-(b). Compared to the localized attacks, here we observe a larger number of trojaned models with low Q scores. However, by setting the Q threshold to $0.3$, we obtain a low false negative rate of $2\%$ with a false positive rate of only $9\%$, and the false negative and false positive rate for all choices of threshold can be seen in~\ref{fig:trojai_FPR}-(b) along with a selection of specific results in~\ref{tab:trojai}.  Prior detection methods require significant side information about the nature of the trigger are generally unable to address an attack such as the Instagram filters, ours is the first to be able to detect them with such high accuracy.  
	
\begin{figure}
\centering
\subfloat[Polygon trigger models]{ \begin{tabular}{||c c c ||} 
 \hline
 Q & FPR & FNR  \\ [0.5ex] 
 \hline\hline
 .297 & .086 & .010  \\
  \hline
 .433 & .018 & .045 \\ 
 \hline
 .492 & .010 & .080 \\ [1ex] 
 \hline
\end{tabular}}
\qquad
\subfloat[Instagram trigger models]{\begin{tabular}{||c c c ||} 
 \hline
 Q & FPR & FNR  \\ [0.5ex] 
 \hline\hline
 .121 & .446 & .047  \\ 
 \hline
 .199 & .243 & .093  \\
 \hline
 .293 & .090 & .163   \\
 \hline
 .370 &  .040 &.197    \\[1ex] 
 \hline
\end{tabular}}
\caption{False positive and negative rates at specific choices of Q threshold from Figure~\ref{fig:trojai_FPR}.  These Q were chosen to attain $1\%$ or $5\%$ for each rate on the Polygon models and $5\%$ or $10\%$ on the Instagram models.}
\label{tab:trojai}
\end{figure}	

	\noindent\textbf{$\blacktriangleright$ Sensitivity to Poisoned Training Data Ratio.} Existing work shows a correlation between the number of the poisoned training data and strength of the attack ~\cite{gao2019strip}. 
	In this section, we study the dependence of our detection results on the proportion of the training data that was poisoned.   
	
\begin{figure}
    \centering
        \subfloat[Polygon trigger models]{\includegraphics[width=.49\columnwidth]{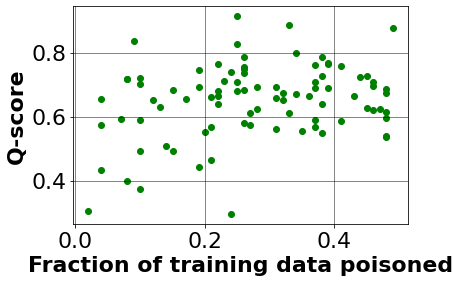}}
        \subfloat[Instagram trigger models]{\includegraphics[width=.49\columnwidth]{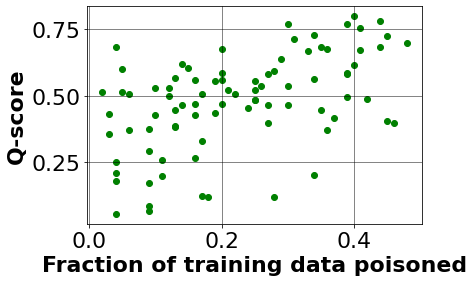}}
	\caption{The Q-scores of the models as a function of the percentage of training data that was poisoned with the trojan trigger. }
	\label{fig:Qvpercent}
\end{figure}

	Figure \ref{fig:Qvpercent} shows the Q-scores of both polygon and Instagram triggered models as a function of the percentage of the training data that was poisoned. There is a positive correlation between the two quantities, with correlation coefficient 0.27 for the polygon models and 0.51 for the Instagram models.  
	This agrees with intuition and prior results that poisoning more data creates a stronger attack.  And it shows that it leaves a more distinct signature in the weights of the final layer of the poisoned network. So a more powerful, reliable trigger will be more easily detected by our method. This forces the attacker to choose between the efficacy of their attack and the ease with which we can detect it.

\subsection{GTSRB}
\label{subsec:GTSRB}
\noindent\textbf{$\blacktriangleright$ The GTSRB Benchmark.} In this part of our analysis, we direct our focus on a single convolutional deep network architecture studied in prior works on trojan attacks~\cite{wang2019neural,javaheripi2020cleann,chen2019deepinspect}. This allows us to perform an in-depth study of different variations of the trojan attack using the same benchmark. The model is trained on the GTSRB dataset which comprises 43 classes of German traffic signs and is a common choice of dataset for trojan research. 

\begin{figure}[h]
    \centering
    \subfloat[Target class 0]{\includegraphics[width=.49\columnwidth]{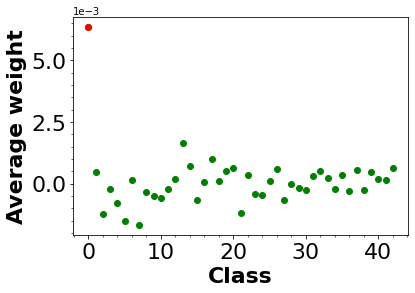}}
    \subfloat[Target class 3]{\includegraphics[width=.49\columnwidth]{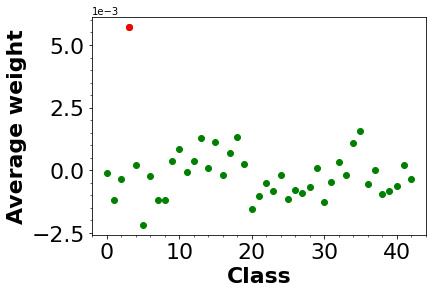}}
    
    \subfloat[Benign]{\includegraphics[width=.49\columnwidth]{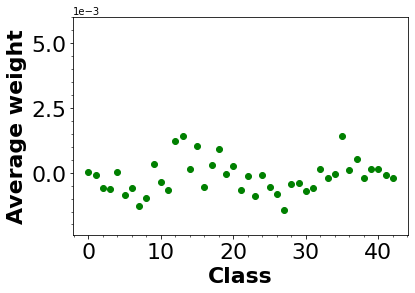}}
    \subfloat[Benign]{\includegraphics[width=.49\columnwidth]{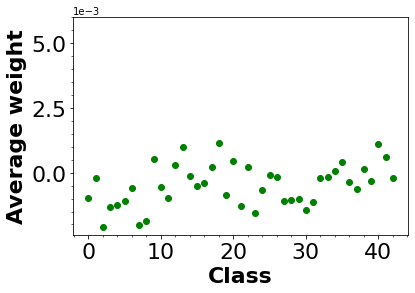}}
    \caption{Average weight per class of four different GTSRB models, two trojaned with different targets and two benign}
        \label{fig:GTSRB_means}
\end{figure}

\noindent\textbf{$\blacktriangleright$ Trojan and Benign Models, Side by Side.} Figure~\ref{fig:GTSRB_means} shows the average row weights of four GTSRB models, two trojaned with different targets and two benign models, all trained under identical conditions except for the choice of target class.  As seen, in each trojaned model, the average weight of the target row is a clear outlier and the benign models show no notable outliers.

\noindent\textbf{$\blacktriangleright$ Sensitivity to Dataset Class Imbalance.} The use of GTSRB dataset allows us to explore a dimension of the problem not present in TrojAI: dataset class imbalance. Many data poisoning attacks, e.g., the BadNet attack, result in an imbalanced training set as they add samples to the target class. This could induce a bias in the network, regardless of trojan behavior. The GTSRB data is already heavily imbalanced, ranging from only 210 training examples for class 0 to 2250 for class 3. We study the effect of imbalance between these two example classes in Figure~\ref{fig:GTSRB_means}. As shown, the weight row corresponding to the target class persists as a clear outlier when either class 0 or class 3 is chosen as the target class. This shows that the trojan signature is independent of the distribution of the training data.

\begin{figure}[h]
    \centering
    \subfloat{\includegraphics[width=.75\columnwidth]{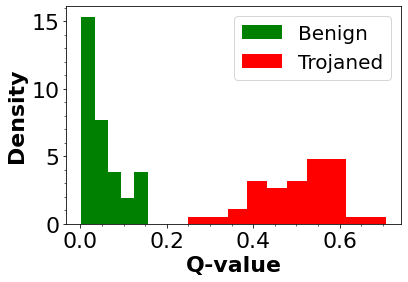}}
    \caption{Normalized histograms of Q-values for 20 benign and 40 trojaned models trained on the GTSRB dataset. Benign and trojaned models are shown with red and green, respectively.}
    \label{fig:GTSRB_Q}
\end{figure}


\noindent\textbf{$\blacktriangleright$ Sensitivity to Attack Parameters.} Finally, we train 20 benign and 40 trojaned models with varying target classes, with the trigger size ranging from $2\times2$ to $12\times12$ pixels in size and located in different corners of the image, and with different proportions of data poisoned, ranging from $10\%$ to $50\%$.  We calculate the Q-score of the average weights for each model and display them in Figure \ref{fig:GTSRB_Q}.  This shows that, in this problem setting, the benign and trojaned models are entirely separable, with no false positives or false negatives, under our detection method.

\subsection{TrojanNN}
    All attacks studied so far are model-agnostic, i.e., the trojan trigger is selected independently from the model and is applied by the training the model, from scratch, on the poisoned dataset. In this section, we study the the TrojanNN attack ~\cite{liu2017trojaning} which constructs the trigger from a pre-trained model.     This sophisticated attack identifies neurons from intermediate layers which have an outsized impact on the output of the pre-trained network. Their algorithm then reverse engineers a trigger which maximizes the activation of those internal neurons.  This attack generally proves more difficult to detect than BadNet and other straightforward data poisoning attacks~\cite{wang2019neural,chen2019deepinspect}.   

\begin{figure}
    \centering
    \subfloat[Square trigger]{\includegraphics[width=.49\columnwidth]{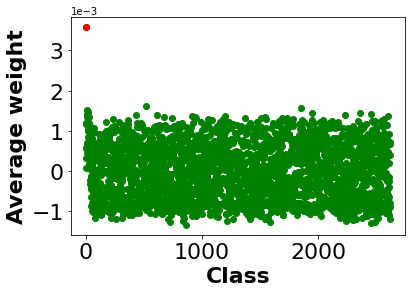}}
    \subfloat[Watermark trigger]{\includegraphics[width=.49\columnwidth]{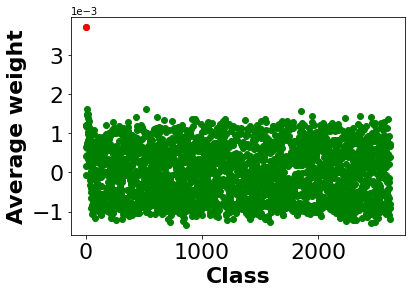}}
    \caption{Average weights per row for models trained on 2000 facial recognition classes, poisoned with the TrojanNN attack.}
    \label{fig:trojann_face}
\end{figure}

\begin{figure}
    \centering
    \subfloat[Attack on $4^{th}$ convolution]{\includegraphics[width=.459\columnwidth]{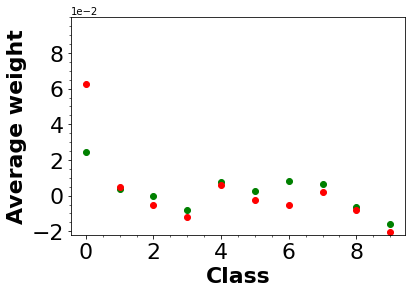}}
    \subfloat[Attack on $1^{st}$ fully-connected]{\includegraphics[width=.49\columnwidth]{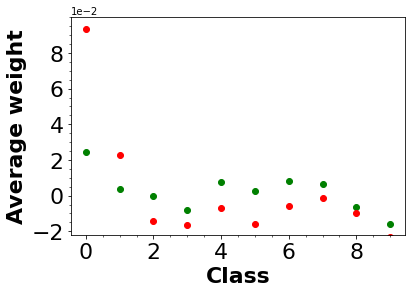}}
    \caption{Average weights of two different implementations of the TrojaNN attack, both with target 0, shown in red, compared to benign analogs, shown in green}
    \label{fig:trojann_speech}
\end{figure}		
    
\noindent\textbf{$\blacktriangleright$ The TrojanNN Benchmark.} 
The creators of the TrojanNN attack provide a series of models poisoned with their attack as well as analogous benign models\footnote{\url{https://github.com/PurduePAML/TrojanNN}}.  We analyze these networks to see if this more subtle attack produces the same weight signature we saw in our previous results.
	
\noindent\textbf{$\blacktriangleright$ Face Recognition Benchmark.} Figure~\ref{fig:trojann_face} compares two examples of a VGGFace network~\cite{parkhi2015deep} poisoned with the TrojanNN attack, both trained on the labeled faces in the wild facial recognition dataset~\cite{LFWTech}.  Both plots show the average weights of each row in the weight matrix of the final layer of the network.  The network in Figures~\ref{fig:trojann_face}-(a), (b) are poisoned with the square and watermark triggers shown in Figure \ref{fig:trojan_triggers}-(b), (c), respectively.  In both models, 0 is the target class and the associated average weight is a clear outlier, so even this more carefully targeted attack, which deliberately acts through a small set of neurons in the intermediate layers of the network, leaves a distinct signature in the weights of the final layer.   

While the two figures look almost identical, they are indeed from separate models with separate triggers and the weights do have different values.  This uniformity is in part due to the fact that the TrojanNN attack is applied on a pre-trained network.  The attack itself is also designed to leave a smaller footprint in the network, as it targets specific neurons, reverse engineers an efficient trigger, and only re-trains a portion of the network.  This makes it all the more notable that the weights of the target row are so drastically exaggerated and thus easily detectable with our analysis.  
	
\noindent\textbf{$\blacktriangleright$ Speech Recognition Benchmark.} Figure \ref{fig:trojann_speech} compares three different networks trained on a speech detection dataset\footnote{\url{https://github.com/pannous/caffe-speech-recognition}}.  The attacks in Figure~\ref{fig:trojann_speech}-(a),~(b) were executed on internal neurons from the fourth convolution and the first fully connected layers, respectively. For both models, the average weight of the target row, row 0 is a clear outlier. Notably, in both cases the average weight of the target row is increased, while the average weight for all other classes is decreased.  
	
The effect here is far more pronounced when the attack is executed on neurons in the fully connected layer than the attack focused on neurons in the convolution layer--the average target weight increased by almost twice as much.  This makes sense in light of the assumptions underlying our analysis.  
The retraining in the TrojaNN algorithm freezes all layers prior to the target neurons--so the action of the trigger is only embedded in the later layers of the network.  Our method relies on the notion that many features of the underlying input will still be present in the later feature representations of the poisoned input.  Executing the TrojanNN attack on neurons from later layers ensures that this assumption holds, as it leaves the feature extraction mechanism of the network largely unchanged and only re-trains the classification layers.     

\begin{figure}[h]
    \centering
    \includegraphics[width=.55\columnwidth]{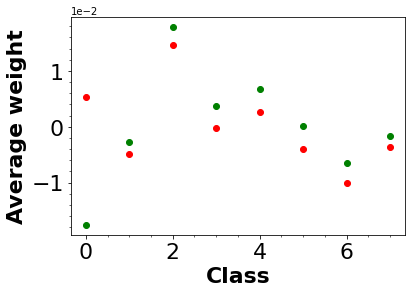}
    \caption{Average weights of the TrojanNN age identification model with target 0, in red, compared with a benign analog, in green.  }
    \label{fig:trojann_age}
\end{figure}

\noindent\textbf{$\blacktriangleright$ Age Detection Benchmark.} Finally, Figure~\ref{fig:trojann_age} compares the average row weights of a benign network and a TrojanNN attacked network trained to perform age classification~\cite{levi2015age}.  Here the target class was 0, so our approach does not successfully identify the trojan. However, we still see the effect characteristic of the other trojan attacks--the average target weight is increased, while other average weights all decreased. Detection here fails because the change in weights is relatively small, compared to the other models we have examined, and because the weight of the target row happened to be abnormally small to begin with. 

\subsection{Adaptive attack}

To study the robustness of this signature, we also implemented an adaptive attack intended to evade our detection mechanism and mask the observed shift in weights.  To this end we added a regularization to the standard cross-entropy loss used in our other experiments.  This regularization penalizes the gap between the average weight of the target row and the average weights of the other rows:
\begin{equation}
    L_{reg} = L_{CE} + \gamma \cdot \big[\E[\mbf{W}_t] - \E[\mbf{W}]\big].
    \label{eq:regloss}
\end{equation}
Where $\E[\mbf{W}]$ denotes the average value of all weights in the final weight matrix and $\gamma$ is a free parameter which controls the strength of the regularization.  We note that this is the loss function used during training, by the adversary, so they know the target, $t$, and can directly regularize the associated statistic.  

\begin{figure}[h]
    \centering
    \subfloat[$\gamma =0$]{\includegraphics[width=.49\columnwidth]{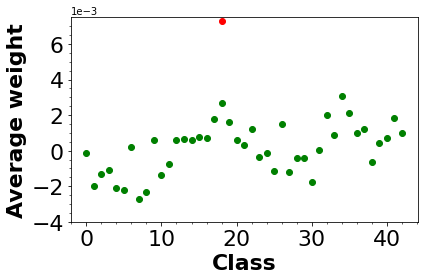}}
    \subfloat[$\gamma=0.005$]{\includegraphics[width=.49\columnwidth]{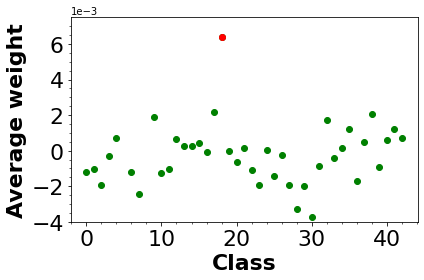}}
    
    \subfloat[$\gamma=0.01$]{\includegraphics[width=.49\columnwidth]{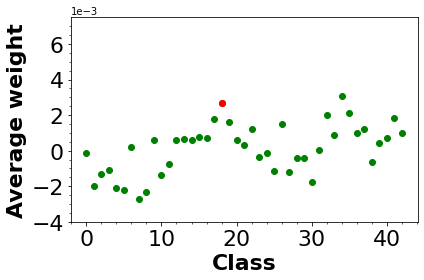}}
    \subfloat[$\gamma=0.05$]{\includegraphics[width=.49\columnwidth]{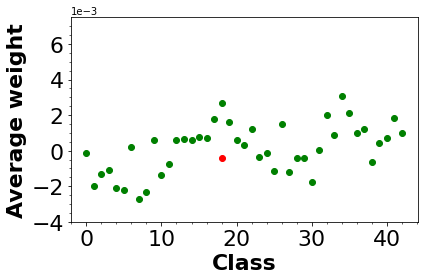}}
    \caption{Average weight per class of for models trained with four different values of the regularization parameter $\gamma$ in Equation~\ref{eq:regloss}}
        \label{fig:regularized_means}
\end{figure}

We trained sets of 10 trojaned models on the GTSRB dataset with this modified loss for a range of values of $\gamma$, with methodology and architecture otherwise identical to that used in Section~\ref{subsec:GTSRB}. Figure~\ref{fig:regularized_means}
shows the average weight of each row of the final weight matrix, with the average of the target row highlighted in red.  This shows that, as $\gamma$ increases, it is effective at regulating the average weight of the target row and bringing it in line with the others, thus masking the signature we analyze.  However, this causes a decrease in both clean accuracy and trigger efficacy: the unregularized models have an accuracy of $.976 \pm .003$ on clean data while the regularized models have accuracy $.958 \pm .005$. And, in the presence of the trojan trigger, the unregularized models classify to the target class at a rate of $.991\pm.003$ while the regularized models only attain $.983\pm .004$.  All these values are averaged over 10 models trained independently with identical settings and show a one standard deviation range.  So the regularization produces a less effective trigger and a less accurate classifier. Alongside this generally inferior performance, the lower accuracy on clean data may itself suffice as an indication that the classifier has been tampered with for architectures and datasets where a standard, attainable accuracy is known.  

	\begin{figure}[h]
	    \centering
	    \subfloat[$\gamma=0$]{\includegraphics[width=.49\columnwidth]{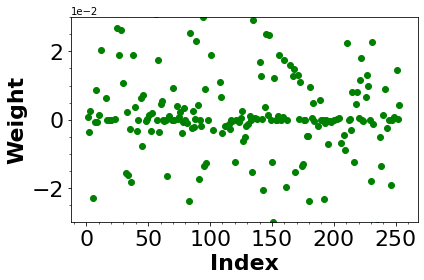}}
	    \subfloat[$\gamma=0.01$]{\includegraphics[width=.49\columnwidth]{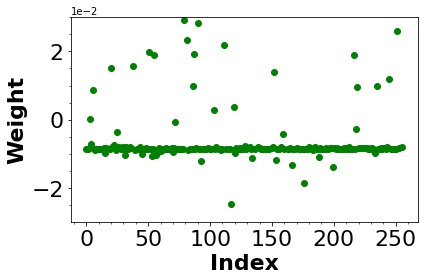}}
	   \caption{The weights of the target rows for a trojaned model trained without regularization ($\gamma=0$) and one trained with regularization, with $\gamma=0.01$. The training process for the two models was identical besides the value of $\gamma$.}
    	\label{fig:regularized_weights}
	\end{figure}

Furthermore, the masking effect is accomplished in a very artificial, easy to detect way.  Figure~\ref{fig:regularized_weights} shows a scatter plot of the weights for the target row in two trojaned models, one without regularization, on the left, and one with $\gamma=0.01$, on the right.  The unregularized version has a relatively diffuse, symmetric distribution of weights with a large concentration near 0.  This is characteristic of the weights for every row, target or otherwise, in every unregularized model, benign or trojaned, that we trained for the GTSRB dataset.  In contrast, on the right in the regularized model the weights are far more concentrated and a large number of weights have been shifted uniformly to a small, negative value.  This occurs only for the target row in trojaned models trained with the regularized loss in Equation~\ref{eq:regloss}. This negative shift in a subset of the weights counteracts the higher average among the rest of the weights.  But the very uniform, artificial shift is easy to detect by eye, and as discussed above, is totally unique to models trojaned in this way.  So while an adversary with full knowledge of our detection mechanism can evade it, they can do so only at the cost of a less accurate classifier, a less effective trigger, and a new signature in the final weight layer.


%% file: 6_related_work.tex
\section{Related Work}

Prior work for trojan detection can be categorized into two different classes based on their execution phase: (1)~offline model inspection methods that aim to find out whether a model has been compromised, and (2)~online input inspection methods that monitor incoming data to the model to discard/correct samples that contain the trojan trigger. Below we review the work in each category in more detail.

\noindent{\bf Model inspection.} Current methods for offline model inspection rely on reverse-engineering the trigger to confirm whether a model has been trojaned or not. Neural Cleanse~\cite{wang2019neural} uses a clean subset of data to solve an optimization problem and extract the potential trojan triggers. It then uses the $L_1$ norm of the generated triggers to decide whether any of them corresponds to a viable trojan. Follow up work~\cite{chen2019deepinspect} proposes to use a conditional GAN to replace the clean dataset and improve the runtime. While these methods achieve good performance on simple triggers, their performance degrades in face of more complex triggers. Furthermore, the computational overhead required by reverse-engineering stage hinders their applicability for users without access to abundant computing resources. 

ABS~\cite{liu2019abs} proposes stimulating neurons and studying the model's behavior to extract trojan triggers. Building upon the idea of examining internal neurons,~\cite{wang2020practical} uses adversarial perturbations as well as random noise to identify trojan signatures. Both of these methods require multiple rounds of forward and backward propagations that make the detection scheme computationally complex.
 Our method is different than the works in this category in that, instead of reverse engineering the trigger, we study the statistics of the last layer's parameters to detect abnormal trojan behavior. This approach enables our method to be universally applicable complex trojan trigger patterns.

\noindent \textbf{Data Inspection.} A line of work in data inspection focus on finding regions of the input image that potentially contain the trojan trigger. This is done by using back-propagation to extract the most influential input regions in classification. Once such regions are extracted, Sentinet~\cite{chou2018sentinet} applies them on a set of benign samples and analyzes the model's change of output to assert if the found region was a trojan trigger. Februus~\cite{doan2019februus} takes a similar approach to find trojan regions and later uses GANs to inpaint the trojan trigger and correct the model's decision. 
STRIP~\cite{gao2019strip} suggests that while injecting noise to benign data significantly varies the predicted class, trojan samples are more robust to such noise patterns. Therefore, they detect trojans by injecting multiple intentional noise patterns and observing the model output prediction. Perhaps the most notable downside of the above works is the heavy computation overhead of backward propagation that hinders their application in latency-sensitive online tasks. 

Analyzing the statistics of benign input samples and identifying outliers has also been investigated in contemporary research. Authors of~\cite{ma2019nic} and ~\cite{chen2018detecting} propose to apply clustering on latent feature maps to detect trojan samples. These methods, however, require access to the model's training dataset including the trojan samples. Authors of~\cite{javaheripi2020cleann} perform sparse recovery to reconstruct the input data and the latent features to remove the effect of the trojan trigger from the clean signal.

While the above methods achieve high detection rates for incoming trojan samples, they are inappropriate for safety critical scenarios where the model should be tested for security and safety compliance prior to deployment.

%% file: 7_conclusion.tex
\section{Conclusion}
We propose the first trojan detection mechanism that requires no access to any data, significant computational resources, or specific knowledge about the type of trojan trigger.  By performing analysis only of the parameters of the final layer of the network it can effectively detect both standard data poisoning attacks and the TrojanN attack before deployment of the network.  This makes our detection mechanism ideally suited for the users with access to limited resources and with security sensitive applications who are most vulnerable to trojan attacks. We hope that the lightweight nature of our method allows it to be applied in conjunction with more complicated methods to create more effective, efficient trojan detection mechanisms.  
